\documentclass[twoside,leqno,twocolumn]{article}  
\usepackage{ltexpprt} 
\usepackage{times}
\usepackage{amsmath}
\usepackage{amsfonts}
\usepackage{amssymb}
\usepackage{graphicx}
\usepackage{url}
\usepackage{epstopdf}%
\usepackage{algorithm}
\usepackage{algorithmic}

\newtheorem{definition}{Definition}

\newenvironment{proof-outline}[1][Proof Outline]{\noindent\textbf{#1.} }{\ \rule{0.5em}{0.5em}}
\DeclareMathOperator*{\argmax}{argmax}

\newcommand{\V}{V}

\newcommand{\node}{v}
\newcommand{\B}{B}
\renewcommand{\P}{P}
\newcommand{\R}{R}
\newcommand{\X}{X}
\newcommand{\Y}{Y}
\newcommand{\Z}{Z}

\newcommand{\U}{U}
\newcommand{\W}{W}
\renewcommand{\S}{S}

\renewcommand{\a}{a}

\newcommand{\x}{x}
\newcommand{\y}{y}
\newcommand{\p}{p}
\newcommand{\s}{s}

\newcommand{\nodevalue}{\a}

\newcommand{\parent}{\mathit{pa}}
\newcommand{\parents}{\mathbf{pa}}
\newcommand{\Parents}{\mathbf{PA}}

\renewcommand{\l}{\ell} 
\newcommand{\mb}{\set{mb}} 
\newcommand{\N}{N} 
\renewcommand{\L}{L} 
\newcommand{\m}{m} 

\newcommand{\C}{\set{C}} 
\newcommand{\fterms}{F} 

\newcommand{\assign}{A} 
\newcommand{\grounds}{\#}

\newcommand{\AVG}{\it{AVG}}

\newcommand{\G}{G}

\newcommand{\J}{J}

\newcommand{\type}{\tau} 
\newcommand{\E}{E} 
\newcommand{\e}{e} 
\newcommand{\f}{f}

\renewcommand{\c}{c}
\renewcommand{\R}{R} 
\newcommand{\A}{A} 
\newcommand{\T}{T} 

\newcommand{\D}{\mathcal{D}} 
\newcommand{\dom}{\mathit{dom}} 
\newcommand{\student}{\mathit{Student}}
\newcommand{\I}{\mathit{I}}
\newcommand{\course}{C}
\newcommand{\prof}{\mathit{Professor}}
\newcommand{\person}{\mathit{Person}}

\newcommand{\age}{\mathit{age}}
\newcommand{\intelligence}{\mathit{intelligence}}
\newcommand{\diff}{\mathit{difficulty}}

\newcommand{\grade}{\mathit{grade}}
\newcommand{\gpa}{\mathit{gpa}}
\newcommand{\jack}{\mathit{Jack}}
\newcommand{\jill}{\mathit{Jill}}

\newcommand{\cmpt}{\mathit{CMPT120}}

\newcommand{\true}{\mathit{T}}
\newcommand{\false}{\mathit{F}}
\newcommand{\join}{\textsc{Join-Frequencies}}
\newcommand{\linus}{\textsc{Linus }}
\newcommand{\foil}{\textsc{Foil }}

\def\set#1{\mathbf{#1}}
\def\bs#1{\boldsymbol{#1}}

\begin{document}

\title{\Large Learning Class-Level Bayes Nets for Relational Data}
\author{Oliver Schulte, Hassan Khosravi, Bahareh Bina, Flavia Moser, Martin Ester\\ \{oschulte, hkhosrav, bb18,fmoser, ester\}@cs.sfu.ca\\ School of Computing Science \\Simon Fraser University \\Vancouver-Burnaby, B.C., Canada }

\date{}

\maketitle


\begin{abstract}Many databases store data in relational format, with different types of entities and information about links between the entities. The field of statistical-relational learning (SRL) has developed a number of new statistical models for such data. In this paper we focus on learning class-level or first-order dependencies, which model the general database statistics over attributes of linked objects and links (e.g., the percentage of A grades given in computer science classes). Class-level statistical relationships are important in themselves, and they support applications like policy making, strategic planning, and query optimization. Most current SRL methods find class-level dependencies, but their main task is to support instance-level predictions about the attributes or links of specific entities. We focus only on class-level prediction, and describe algorithms for learning class-level models that are orders of magnitude faster for this task. 
Our algorithms learn Bayes nets with relational structure, leveraging the efficiency of single-table nonrelational Bayes net learners.  An evaluation of our methods on three data sets shows that they are computationally feasible for realistic table sizes, and that the learned structures represent the statistical information in the databases well. After learning compiles the database statistics into a Bayes net, querying these statistics via Bayes net inference is faster than with SQL queries, and does not depend on the size of the database.
\end{abstract}

\section{Introduction}

Many real-world applications store data in relational format, with different tables for entities and their links. Standard machine learning techniques are applied to data stored in a single table, that is, in nonrelational, propositional or ``flat" format \cite{Mitchell1997}. The field of statistical-relational learning (SRL) aims to extend machine learning algorithms to relational data \cite{getoor-intro,deRaedt08}. One of the major machine learning tasks is to use data to build a {\em generative statistical model} for the variables in an application domain \cite{getoor-intro}.
In the single-table learning setting, the goal is often to represent predictive dependencies between the attributes of a single individual (e.g., between the intelligence and ranking of a student). In the SRL setting, the goal is often to represent, in addition, dependencies between attributes of different individuals that are related or linked to each other (e.g., between the intelligence of a student and the difficulty of a course given that the student is registered in the course). 

\subsection*{Task Description: Modelling Class-Level Dependencies} 
Many SRL models distinguish two different levels, a class or type dependency model $\G_{M}$ and an instance dependency model $\G_{I}$ \cite{Getoor2007c,Kersting2007,bib:jensen-chapter}. 
In a graphical SRL model, the nodes in the instance dependency model represent attributes of individuals or relationships. The nodes in the class dependency model correspond to attributes of the tables. The use of the term ``class'' here is unrelated to classification; \cite{Koller1997} views it as analogous to the concept of class in object-oriented programming. An example of a class-level probabilistic dependency is ``among students with high intelligence, the rate of GPA = 4.0 is 80\%''. An instance-level prediction would be ``given that Jack is highly intelligent, the probability that his GPA is 4.0 is 80\%''. Thus class-level dependencies are concerned with the {\em rates} at which events occur, or at which properties hold within a class, whereas instance-level dependencies are concerned with specific events or the properties of specific entities \cite{Halpern90,Bacchus90}. 
In terms of predicate logic, the class-level model features terms that involve first-order variables (e.g., $\it{intelligence}(\S)$, where $\S$ is a 
variable ranging over a domain like $\it{Students}$), whereas the instance-level graph features terms that involve constants (e.g., $\it{intelligence}(\it{jack})$ where $\it{jack}$ is a constant that denotes a particular student) \cite{deRaedt08,Kersting2007,Domingos07}. 

Typically, SRL systems instantiate a class-level model $\G_{M}$ with the specific entities, their attributes and their relationships in a given database to obtain an instance dependency graph $\G_{I}$. An important purpose of the instance graph is to support predictions about the attributes of individual entities.
A key issue for making such predictions is the {\em combining problem}: how to combine information from different related entities to predict a property of a target entity. 
Current SRL model construction algorithms learn class-level dependencies and instance-level predictions at the same time. What is new about our approach is that we focus on class-level variables only rather than making predictions about individual entities. 
We apply Bayes net technology to design new algorithms especially for learning class-level dependencies. In experiments our algorithms run at two orders of magnitude faster than a benchmark SRL method.
Our models thus trade-off tractability of learning with the ability to answer queries about individual entities.

\paragraph{Applications.} Examples of applications that provide motivation for the class-level statistical models include the following. 

\begin{enumerate}
\item {\em Policy making and strategic planning.} A university administrator may wish to know which program characteristics attract high-ranking students in general, rather than predict the rank of a specific student in a specific program. 
\item {\em Query optimization} is one of the applications of SRL where a statistical model predicts a probability for given table join conditions that can be used to infer the size of the join result \cite{Getoor2001}. Estimating join sizes is a key problem for database query optimization.
In queries that involve several tables being joined together, the ideal scenario is to have smaller intermediate joins \cite{McMahan2004}. A class-level statistical model may be used for estimating frequency counts in the database to select smaller joins, and so to optimize the speed of answering queries by taking more efficient intermediate steps.
For example, suppose we wish to predict the size of the join of a $\it{Student}$ table with a $\it{Registered}$ table that records which students are registered in which courses, where the selection condition is that the student have  high intelligence. In a logical query language like the domain relational calculus \cite{Ullman1982}, this join would be expressed by the conjunctive formula $\it{Registered}(\S,\course), \it{intelligence}(\S) = \it{high}$. A query to a Join Bayes Net can be used to estimate the frequency with which this conjunction is true in the database, which immediately translates into an estimate of the size of the join that corresponds to the conjunction. 
The join conditions often do not involve specific individuals.
\item {\em Private Data.} In some domains, information about individuals is protected due to confidentiality concerns. For example, \cite{Frank2009} analyzes a database of police crime records. The database is anonymized and it would be unethical for the data mining researcher to try and predict which crimes have been committed by which individuals. However, it is appropriate and important to look for general risk factors associated with crimes, for example spatial patterns \cite{Frank2009}. 
Under the heading of privacy-preserving data mining, researchers have devoted much attention to the problem of discovering class-level patterns without compromising sensitive information about individuals \cite{Verykios2004}.
\end{enumerate}

\paragraph{Approach.} 

We apply Bayes nets (BNs) to model class-level dependencies 
between attributes that appear in separate tables. Bayes nets \cite{Pearl1988} have been one of the most widely studied and applied generative model classes. A BN is a directed acyclic graph whose nodes represent random variables and whose edges represent direct statistical associations.  Our class-level Bayes nets contain nodes that correspond 
to the descriptive attributes of the database tables, plus Boolean nodes that indicate the presence of a relationship; we refer to these as Join Bayes nets (JBNs). 
To apply machine learning algorithms to learn the structure of a Join Bayes Net from a database, we need to define an {\em empirical database distribution} over values for the class-level nodes that is based on the frequencies of events in the database. In a logical setting, the question is how to assign database frequencies for a conjunction of atomic statements such as $\it{intelligence}(\S) = \it{high}$, $\it{Registered}(\S,\course)$, $\it{difficulty}(\course) = \it{high}$. We follow the definition that was established in fundamental AI research concerning the combination of logic and statistics, especially the classic work of Halpern and Bacchus 
\cite{Halpern90,Bacchus90}. 
This research generalized the concept of single-table frequencies to the relational domain: the frequency of a first-order formula in a relational database is the number of instantiations of the variables in the formula that satisfy the formula in the database, divided by the number of all possible instantiations. In the example above, the instantiation frequency would be the number of student-course pairs where the student is highly intelligent, the course is highly difficult, and the student is registered in the course, divided by all possible student-course pairs. In terms of table joins, the instantiation frequency is the number of tuples in the join that corresponds to the conjunction, divided by the maximum size of the join. 

Our learn-and-join algorithm aims to find a model of the database distribution. It upgrades a single table BN learner, which can be chosen by the user, to carry out relational learning
by decomposing the learning problem for the entire database into learning problems for smaller tables. The basic idea is to apply the BN learner repeatedly to tables and join tables from the database, and merge the resulting graphs into a single graphical model for the entire database. 
This algorithm treats the single-table BN learner as a module within the relational structure learning system. This means that only minimal work is required to build a statistical-relational JBN learner from a single-table BN learner.

The main algorithmic problem in parameter estimation (CP-table estimation) for Join Bayes nets is counting the number of satisfying instantiations (groundings) of a first-order formula in a database. We show how the recent virtual join algorithm of \cite{Khosravi2009} can be applied to solve this problem efficiently. 
The virtual join algorithm is an efficient algorithm designed to compute database instantiation frequencies that involve non-existent links.

Our experiments provide evidence that the learn and  join algorithm leverages the efficiency, scalability and reliability of single-table BN learning into efficiency, scalability and reliability for statistical-relational learning. We benchmark the computational performance of our algorithm against structure learning for Markov Logic Networks (MLNs), one of the most prominent statistical-relational formalisms \cite{Domingos07}. In our experiments on small datasets, the run-time of the learn-and-join algorithm is about 20 times faster than the state-of-the art Alchemy program  \cite{Domingos07} for learning MLN structure. On medium-size datasets, such as the Financial database from the PKDD 1999 cup, Alchemy does not return a result given our system resources, whereas the learn-and-join algorithm produces a Join Bayes net model within less than 10 min. To evaluate the learned structures, we apply BN inference algorithms to predict relational frequencies and compare them with gold-standard frequencies computed via SQL queries. The learned Join Bayes nets predict the database frequencies very well. Our experiments on prediction take advantage of the fact that because JBNs use the standard Bayes net format, class-level queries can be answered by standard BN inference algorithms that are used ``as is''. 
Other SRL formalisms focus on instance-level inference, and do not support class-level inference, at least in their current implementations. Our datasets and code are available for ftp download from ftp://ftp.fas.sfu.ca/pub/cs/oschulte.


\paragraph{Paper Organization.}
As statistical-relational learning is a complex subject and a variety of approaches have been proposed, we review related work in some detail. A preliminary section introduces Bayes nets and predicate logic. We formally define Join Bayes nets and the instantiation frequency distribution that they model. The main part of the paper describes structure and parameter learning algorithms for Join Bayes nets. We evaluate the algorithms on one synthetic dataset and two public domain datasets (MovieLens and Financial), examining the learning runtimes and the predictive performance of the learned models. 

\paragraph{Contributions.}
The main contributions of our work are as follows.

(1) A new type of first-order Bayes net for modelling the database distribution over attributes and relationships, which is defined by applying classic AI work on probability and logic.

(2) New efficient algorithms for learning the structure and parameters for the first-order Bayes net models.

(3) An evaluation of the algorithms on  three relational databases, and an evaluation of the predictive performance of the learned structures.


\section{Related Work} \label{sec:related}

A preliminary version of our results appeared in the proceedings of the STRUCK and GKR workshops at IJCAI 2009.
There is much AI research on the theoretical foundations of the  database distribution, and on representing and reasoning with instantiation frequencies in an axiomatic framework based on theorem proving \cite{Halpern90,Bacchus90}. Our approach utilizes graphical models for learning and inference with the database distribution rather than a logical calculus.
For inference, our approach appears to be the first that utilizes standard BN inference algorithm to carry out probabilistic reasoning about database frequencies. For learning, our work appears to be the first to use Bayes nets to learn a statistical model of the database distribution. 

\subsection*{Other SRL formalisms.} Various formalisms have been proposed for combining logic with graphical models, such as Bayes Logic Networks (BLNs) \cite{Kersting2007}, parametrized belief networks~\cite{Poole2003}, object-oriented Bayes nets \cite{Koller1997}, Probabilistic Relational Model (PRMs) \cite{Getoor2007c}, and Markov Logic Networks \cite{Domingos07}. We summarize the main points of comparison. General SRL overviews are provided in \cite{Kersting2007,getoor-intro,deRaedt08}. The main difference between JBNs and other SRL formalisms is their semantics. The semantics of JBNs is specified at the class level without reference to an instance-level model; a JBN models the database distribution defined by the instantiation frequencies.

Syntactically, Join Bayes Nets are very similar to parametrized belief networks and to BLNs. BLNs require the specification of combining rules, a standard concept in Bayes net design, to solve the combining problem. For instance, the class-level model may specify the probability that a student is highly intelligent given that she obtained an A in a difficult course. A set of grades thus translates into a multiset of probabilities, which the combining rule translates into a single probability. Essentially, a JBN is a BLN without combining rules. 

The key difference between PRMs and BLNs is that PRMs use aggregation functions (like average), not combining rules \cite{deRaedt08}. 
\cite[Sec.10.4.3]{Kersting2007} shows how aggregate functions can be added to BLNs; the same construction works for adding them to JBNs.  Object-oriented Bayes nets use aggregation like PRMs, and have special constructs for capturing class hierarchies; otherwise their expressive power is similar to BLNs and JBNs.

Markov Logic Networks combine ideas from {\em undirected} graphical models with a logical representation. An MLN is a set of formulas, with a weight assigned to each. MLNs are like JBNs, and unlike BLNs and PRMs, in that an MLN is complete without specifying a combining rule or aggregation function. They are like BLNs and PRMs, and unlike JBNs, in that their semantics is specified in terms of an instance-level network whose nodes contain ground formulas (e.g., $\it{age}(\it{jack}) = 40$). 
Instance-level prediction is defined by using the log-linear form of Markov random fields \cite[Sec.12.4]{Domingos07}. 


%


\subsection*{Inference and Learning}
Because JBNs use the Bayes net format, class-level probabilistic queries that involve first-order variables and no constants can be answered by standard efficient Bayes nets inference algorithms, which are used ``as is''. This can be seen as an instance of lifted or first-order probabilistic inference, a topic that has received a good deal of attention recently \cite{Poole2003,Braz2007}. 
Other SRL formalisms focus on instance-level queries that involve constants only, and do not support class-level inference, at least in their current implementations. 

The most common approach to SRL structure learning is to use the instance-level structure $\G_{I}$ to assign a likelihood to a given database $\D$. This likelihood is the counterpart to the likelihood of a  sample given a statistical model in single table learning. Learning searches for a parametrized model $\G_{M}$ whose instance-level graph $\G_{I}$ assigns a maximum likelihood to the given database $\D$, where typically the likelihood is balanced with a complexity penalty to prevent overfitting \cite{Getoor2007c,Kersting2007}. This approach is a conceptually elegant way to learn class-level dependencies and instance-level predictions at the same time. JBNs separate the task of finding generic class-level dependencies from the task of predicting the attributes of individuals. Two key advantages of learning directed graphical models such as Bayes nets at the class-level only include the following.

(1) Class-level learning avoids the problem that the instantiated model may contain cycles, even if the class-level model does not. 
For example, suppose that the class-level model indicates that if a student $\S_1$ is friends with another student $\S_2$, then their smoking habits are likely to be similar, so $\it{Smokes}(\S_{1})$  predicts $\it{Smokes}(\S_{2})$. Now if in the database we have a situation where $a$ is friends with $b$, and $b$ with $c$, and $c$ with $a$, the instance level model would contain a cycle $\it{Smokes}(a) \rightarrow \it{Smokes}(b) \rightarrow \it{Smokes}(c) \rightarrow \it{Smokes}(a)$
\cite{Getoor2007c,bib:jensen-chapter}. If learning is defined in terms of the instance model, cycles cause difficulties because concepts and algorithms for directed acyclic models no longer apply. In particular, the likelihood of a database that measures data fit is no longer defined. These difficulties have led researchers to conclude that ``the acyclicity constraints of directed models severely limit their applicability to relational data'' \cite{bib:jensen-chapter} (see also \cite{Domingos07,Taskar2002}). 

(2) Other directed SRL formalisms require extra structure to solve the combining problem for instance-level predictions, such as aggregation functions  or combining rules. 
While the extra structure increases the representational power of the model, it also leads to considerable complications in learning.

\section{Preliminaries}
We combine the concepts of graphical models from machine learning, relational schemas from database theory, and conjunctions of literals from first-order logic. This section introduces notations and definitions for these background subjects.

\subsection{Bayes nets}\label{sec:bns}

A random variable is a pair $\X = \langle \dom(\X), P_{\X} rangle$ where $\dom(\X)$ is a set of possible values for $\X$ called the \textbf{domain} of $\X$ and $P_{\X}: \dom(\X) \rightarrow [0,1]$ is a probability distribution over these values. For simplicity we assume in  this paper that all random variables have finite domains (i.e., discrete or categorical variables).
Generic values in the domain are denoted by lowercase letters like $\x$ and $a$. 
An \textbf{atomic assignment} assigns a value $\X = x$ to random variable $\x$, where $\x \in \dom(\X)$. A \textbf{joint distribution} $P$ assigns a probability to each conjunction of atomic assignments; we write $P(\X_1 = \x_1,...,\X_n = \x_n) = p$, sometimes abbreviated as $P(\x_1,...,\x_n) = p$. To compactly refer to a set of variables like $\{\X_1,..\X_n\}$ and an assignment of values $\x_1,..,\x_n$, we use boldface $\set{X}$ and $\set{x}$. If $\set{X}$ and $\set{Y}$ are sets of variables with $P(\set{Y} = \set{y}) > 0$, the \textbf{conditional probability} $P(\set{X}=\set{x}|\set{Y} =\set{y})$ is defined as $P(\set{X}=\set{x},\set{Y} =\set{y})/P(\set{\Y} = \set{\y})$.

We employ notation and terminology from \cite{Pearl1988,Spirtes2000} for a Bayesian Network. A {\bf Bayes net structure} is a directed acyclic graph (DAG) $G$, whose nodes comprise a set of random variables denoted by $\V$. When discussing a Bayes net, we refer interchangeably to its nodes or its variables. The parents of a node $\X$ in graph $\G$ are denoted by $\Parents_{\X}^{\G}$, and an assignment of values to the parents is denoted as $\parents_{\X}^{\G}$. 
When there is  no risk of confusion, we often simply write $\parents_{\X}$.
A Bayes net is a pair 
$\langle{G,\theta_G}rangle$ where $\theta_G$ is a set of parameter values that specify the  probability distributions of children conditional on instantiations of their parents, i.e. all conditional probabilities of the form $P(\X=\x|\parents^{\G}_{\X})$. These conditional probabilities are specified in a \textbf{conditional probability table} (CP-table) for variable $\X$. 
A BN $\langle{G,\theta_G}rangle$ defines a joint probability distribution over $\V = \{\node_1,..,\node_n\}$ according to the formula 
\begin{equation} \notag
P(\set{\node} = \set{\nodevalue}) = \prod_{i=1}^{n} P(v_{i} = a_{i}|\parents_{i} = a_{\parents_{i}})
\label{eq:bayes-factor}
\end{equation}
where $a_{i}$ is the value of node $v_{i}$ specified in assignment $\nodevalue$, the term $a_{\parents_{i}}$ denotes the assignment of a value to each parent of $v_{i}$ specified in assignment $\nodevalue$, and $ P(v_{i} = a_{i}|\parents_{i} = a_{\parents_{i}})$ is the corresponding CP-table entry.
Thus the joint probability of an assignment is obtained by multiplying the conditional probabilities of each node value assignment given its parent value assignments.

\subsection{Relational Schemas and First-Order Formulas}

We begin with a standard \textbf{relational schema} containing a set of tables, each with key fields, 
descriptive attributes, and foreign key pointers. A \textbf{database instance} specifies the tuples contained in the tables of a given database schema.
Table ref{table:university-schema} shows a relational schema for a database related to a university (cf.~\cite{Getoor2007c}), and Figure ref{fig:university-tables} displays a small database instance for this schema.
A \textbf{table join} of two or more tables contains the rows in the Cartesian products of the tables whose values match on common fields.

 \begin{table}[tbp] \centering
{\small
\begin{tabular}
[c]{|l|}\hline
$\student$(\underline{$student\_id$}, $\intelligence$, $ranking$)\\
$\it{Course}$(\underline{$\it{course}\_id$}, $\diff$, $rating$)\\ 
$\prof$ (\underline{$professor\_id$}, $teaching\_ability$, $popularity$)\\
$reg$ (\underline{$student\_id$, $\it{course}\_id$}, $grade$, $satisfaction$)\\
$ra$ (\underline{$student\_id$, $prof\_id$}, $salary$, $capability$)\\
\hline
\end{tabular}
}
\caption{A relational schema for a university domain. Key fields are underlined. 
\label{table:university-schema}} 
\end{table}

\begin{figure}[htbp] 
   \centering
   \includegraphics[width=3.5in]{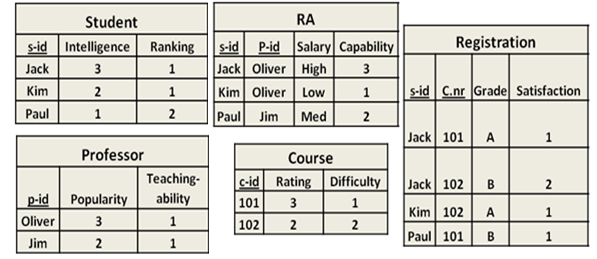} 
  \caption{A database instance for the schema in Table ref{table:university-schema}.}
   \label{fig:university-tables}
\end{figure}

If the schema is derived 
from an entity-relationship model (ER model) \cite[Ch.2.2]{Ullman1982}, the tables in the relational schema can be divided into {\em entity tables} and {\em relationship tables.} Our algorithms generalize to any data model that can be translated into logical vocabulary based on first-order logic, which is the case for an ER model. The entity types of Schema~ref{table:university-schema} are students, courses and professors. There are two relationship tables: $\it{Registration}$ records courses taken by each student and the grade and satisfaction achieved, while $ra$ records research assistantship contracts between students and professors.

In our university example, there are two entity tables: a $\student$ table and a $\course$ table.  There is one relationship table $reg$ with foreign key pointers to the $\student$ and $\course$ tables whose tuples indicate which students have registered in which courses. 
Intuitively, an entity table corresponds to a type of entity, and a relationship table represents a relation between entity types.

It is well known that a relational schema can be translated into \cite{Ullman1982}. 
We follow logic-based approaches to SRL that use logic as a 
rigorous and expressive 
formalism for representing regularities found in the data. Specifically, we  use first-order logic with typed variables and function symbols, as in \cite{Domingos07,Poole2003}. 
 This formalism is rich enough to represent the constraints of an ER schema via the following translation: Entity sets correspond to types, descriptive attributes to functions, relationship tables to predicates, and foreign key constraints to type constraints on the arguments of relationship predicates. 
 
 Formulas in our syntax are constructed using three types of symbols: constants, variables, and functions. Sometimes we refer to logical variables as first-order variables, to distinguish them from random variables. A \textbf{type} or domain is a set of constants. In the case of an entity type, the constants correspond to primary keys in an entity table. Each constant and variable is assigned to a type, and so are the arguments (inputs) and the values (outputs) of functions. A predicate is a function whose values are the special truth values constants $\true,\false$. 
 \cite{Domingos07} discusses the appropriate use of first-order logic for SRL. 
 While our logical syntax is standard, we give the definition in detail, to clarify the space of statistical patterns that our learning algorithms can be applied to, and to give a rigorous definition of the database frequency, which is key to our learning method. Table~ref{table:vocab} defines the logical vocabulary.
 and illustrates the correspondence to an ER schema.
%
\begin{table*}
{\footnotesize \begin{tabular}{|l|l|l|l|}
\hline
Description & Notation & Comment/Definition & Example \\ \hline\hline
Constants & $T,F,\bot ,a_{1},a_{2},\ldots $ & $\bot $ for "undefined" & $%
\it{jack,101,A,B},1,2,3$ \\ \hline
Types & $\type_{1}, \type_{2}, \ldots $ &  list of types & $\it{Student},\it{A,B,C,D}$ \\ \hline
Functions & $f_{1},f_{2},\ldots $ & 
\begin{tabular}{l}
number of arguments = \\ 
arity of $f$%
\end{tabular}
& 
\begin{tabular}{l}
$\it{intelligence,difficulty,grade}$ \\ 
$\it{Student,Registration}$%
\end{tabular}
\\ \hline
Function Domain & $dom(f) = \tau_{i}$ & The value domain of $f$ & 
\begin{tabular}{l}
$\dom(\it{intelligence})=\{1,2,3\}$ \\ 
$\dom(\it{Student})=\{T,F\}$%
\end{tabular}
\\ \hline
Predicate & $P,E,R,$ etc. & A function with domain $\{T,F\}$ & $%
\it{Student,Course,Registration}$ \\ \hline
Relationship & $R,R^{\prime },$ etc. & A predicate of arity greater than 1 & 
$\it{Registration}$ \\ \hline
(Entity) Types & $E_{1},E_{2},\ldots $ & \begin{tabular}{l}
A set of unary
predicates \\ 
that are also types
\end{tabular}
 & $Student,Course$ \\ \hline
(Typed) Variables & $X_{1}^{\type},X_{2}^{\type},\ldots $ & A list of variables of
type $\type$ & 
\begin{tabular}{l}
$S_{1}^{Student},S_{2}^{Student},C_{{}}^{Course}$ \\ 
or simply $S_{1},S_{2},C$%
\end{tabular}
\\ \hline
Argument Type & $\tau (f)=(E_{1},\ldots ,E_{n})$ & A list of argument types
for $f$. & 
\begin{tabular}{l}
$\tau (Registration)=$ \\ 
$(Student,Course)$%
\end{tabular}
\\ \hline
Descriptive Attribute & $f_{1}^{P},f_{2}^{P},\ldots $ & \begin{tabular}{l}
Functions associated
with a predicate $\P$ \\ 
$\type(\f^{\P}) = \type(\P)$%
\end{tabular}

 & $\it{intelligence},\it{difficulty},\it{grade}$ \\ \hline
\end{tabular}%
}
\caption{Definition of Logical Vocabulary with typed variables and function symbols. The examples refer to the database instance of Figure~ref{fig:university-tables}.\label{table:vocab}}
\end{table*}
A \textbf{term} $\theta$ is any expression that denotes a single object; the notation $\boldsymbol{\theta}$ denotes a vector or list of terms. 
If $\P$ is a predicate, we sometimes write $\P(\boldsymbol{\theta})$ for $\P(\boldsymbol{\theta}) = \true$ and $\neg \P(\boldsymbol{\theta})$ for $\P(\boldsymbol{\theta}) = \false$.
Terms are recursively constructed as follows.

\begin{enumerate}
\item A constant or variable is a term.
\item Let $f$ be a function term with argument type $\type(f) = (\tau_{1},\ldots,\tau_{n})$. A list of terms $\theta_{1},\ldots,\theta_{n}$ matches $\type(f)$ if each $\theta_{i}$ is of type $\tau_{i}$. If $\boldsymbol{\theta}$ matches the argument type of $f$, then the expression $\f(\boldsymbol{\theta})$ is a \textbf{function term} whose type is $\dom(\f)$, the value type of $f$.
\end{enumerate}

An \textbf{atom} is an equation of the form $\theta = \theta'$ where the  types of $\theta$ and $\theta'$ match.
A \textbf{negative literal} is an atom of the form $\P(\bs{\theta}) = \false$; all other atoms are \textbf{positive literals}. 
%
The formulas we consider are \textbf{conjunctions of literals}, or for short just conjunctions. We use the Prolog-style notation $\L_{1}, \ldots, \L_{n}$ for $\L_{1} \wedge \cdots \wedge \L_{n}$, and vector notation $\set{\L},\set{\C}$ for conjunctions of literals. A term (literal) is \textbf{ground} if it contains no variables; otherwise the term (literal) is \textbf{open}.  If $\fterms = \{\f_{1},\ldots,\f_{n}\}$ is a finite set of open function terms $\fterms$, an \textbf{assignment} to $\fterms$ is
a conjunction of the form $\assign = (\f_{1} = \a_{1},\ldots, \f_{n} = \a_{n})$, where each $\a_{i}$ is a constant.
A \textbf{relationship literal} is a literal with two or more variables. 

A database instance $\D$ (possible world, Herbrand interpretation) assigns a denotation constant to each ground function term $\f(\set{a})$
which we denote by

\[
[f(\set{a})]_{\D}.
\]
The value of descriptive relationship attributes is not defined for tuples that are not linked by the relationship. (For example, $\it{grade}(\it{kim},\it{101})$ is not well defined in the database instance of Figure~ref{fig:university-tables} since $\it{Registered}((\it{kim},\it{101})$  is false in $\D$.) In this case, we assign the descriptive attribute the special value $\bot$ for ``undefined''. 
The general constraint is that for any descriptive attribute $\f^{\P}$ of a predicate $\P$,

\begin{equation} \notag
[f^{\P}(\set{a})]_{\D} = \bot \iff [\P(\set{a})]_{\D} = \false. \label{eq:undefined}
\end{equation}

For a ground literal $\L$, we write $\D \models \L$ if $\L$ evaluates as true in $\D$, and $\D \not\models \L$ otherwise. If $\P$ is a predicate, we sometimes write $\D \models \P(\set{a})$ for $\D \models (\P(\set{a}) = \true)$ and $\D \models \neg \P(\set{a})$ for $\D \models (\P(\set{a}) = \false)$. A ground conjunction $\C$ evaluates to true  just in case each of its literals evaluate to true.
We make the {\em unique names assumption} that distinct constants denote different objects, so 
\[
\D \models (\theta = \theta') \iff  [\theta]_{\D} = [\theta']_{\D}.
\]
If $\E$ is an entity type, the domain of $\E$ is the set of constants satisfying $\E$ (primary keys in the $\E$ table). 
The \textbf{domain of a variable $\X^{\T}$} of type $\T$ is the same as the domain of the type, so $\dom_{\D}(\X^{\T}) = \dom_{\D}(\T)$. 
%
%
A \textbf{grounding} $\gamma$ for a set of variables $\X_{1},\ldots,\X_{k}$ assigns a constant of the right type to each variable $\X_{i}$
(i.e., $\gamma(\X_{i}) \in \dom_{\D}(\X_{i}))$. 
If $\gamma$ is a grounding for all variables that occur in a conjunction $\C$, we write $\gamma \C$ for the result of replacing each occurrence of a variable $\X$ in $\C$ by the constant $\gamma(\X)$. 
The number of groundings that satisfy a conjunction $\C$ in $\D$ is defined as

\[
\grounds_{\D} =|\{\gamma: \D \models \gamma \C\}|
\] 
where $\gamma$ is any grounding for the variables in $\C$, and $|S|$ denotes the cardinality of a set $\S$. 

The ratio of the number of groundings that satisfy $\C$, over the number of possible groundings is called the \textbf{instantiation frequency} or the \textbf{database frequency} of $\C$. Formally we define

\begin{equation}
P_{\D}(\C)= 
\frac{\grounds_{\D}(\C)}{|\dom_{\D}(\X_{1})| \times \cdots \times |\dom_{\D}(\X_{k})|} \label{eq:prob}
\end{equation}

where $\X_{1},\ldots,\X_{k}, k>0$, is the set of variables that occur in $\C$. 

\subsection*{Discussion.} When all function terms contain the same single variable $\X$, Equation~\eqref{eq:prob} reduces to the standard definition of the single-table frequency of events. For example, $P_{\D}(\it{intelligence}(\S) = 3)$ is the ratio of students with an intelligence level of 3, over the number of all students. Halpern gave a definition of the frequency with which a first-order formula holds in a given database \cite[Sec.2]{Halpern90}, which assumes a distribution $\mu$ over the domain of each type. The instantiation frequency~\eqref{eq:prob} is a special case of his with a uniform distribution $\mu$ over the elements of each domain \cite[fn.1]{Halpern90}. As Halpern explains, an intuitive interpretation of this definition is that it corresponds to generic regularities or random individuals, such as ``the probability that a randomly chosen bird will fly is greater than .9''. The conditional instantiation frequency plays an important role for discriminative learning in Inductive Logic Programming (ILP). For instance, the classic \foil system generalizes entropy-based decision tree learning to first-order rules, where the goal is to predict a class label $\l$.
\foil uses the conditional instantiation frequency $\P_{\D}(\l|\C)$
to define the entropy of the empirical class distribution conditional on the body $\C$ of a rule.

\subsection*{Examples.} The examples refer to the database instance $\D$ of Figure~ref{fig:university-tables}.
The entity types are $\it{Student},\it{Course},\it{Professor}$.
For each entity type we introduce one variable $\S,\course,\P$. The constants of type $\it{Student}$ are $\{\it{jack},\it{kim},\it{paul}\}$. 
True ground literals include $\it{intelligence}(\it{jack})=3$ and $\neg \it{Registered}(\it{kim},101)$. 
Table~ref{table:literals} shows various open literals and their frequency in database $\D$ as derived from the number of true groundings. 

\begin{table}[htdp]
\begin{center}\begin{tabular}{|c|c|c|}\hline Literal(s) $L$& $\grounds_{\D}(\L)$ & $P_{\D}(\L)$
 \\\hline 
 $\it{intelligence}(\S)=1$ & 1 & 1/3 
 \\\hline $\neg \it{intelligence}(\S) = 1$ & 2 & 2/3 
 \\\hline $\it{difficulty}(\course) = 2$ & 1 & 1/2 
 \\\hline $\it{Registered}(\S,\course)$ & 4 & 4/6 
 \\\hline $\neg \it{Registered}(\S,\course)$ & 2 & 2/6
 \\\hline $\it{grade}(\S,\course)=B$ & 2 & 2/6
 \\\hline $\it{grade}(\S,\course) = B,$ $\it{salary}(\S,\P)=\it{hi}$ & 1 & 1/12 \\\hline
 \end{tabular} \caption{Examples of open literals and their frequency in the database instance of Figure~ref{fig:university-tables}. The bottom four lines show relationship literals.\label{table:literals}}
\end{center}
\end{table}

\section{Structure Learning for Join Bayes Nets} \label{sec:jbn}



We define the class of Bayes net models that we use to model the database distribution and present a structure learning algorithm.

\begin{definition} \label{def:jbn} Let $\D$ be a database with associated logical vocabulary $vocab$. 
A Join Bayes Net (JBN) structure for $\D$ is a directed acyclic graph whose nodes are a finite set $\{\f_{1}(\bs{\theta_{1}}),\ldots,\f_{n}(\bs{\theta_{n}})\}$ of open function terms. The domain of a node $v_{i} = \f_{i}(\bs{\theta_{i}})$ is the range of $\f_{i}$ (the set of possible output values for $\f_{i}$).
\end{definition}
%
%
Figure~ref{fig:university-JBN} shows an example of a Join Bayes net. We also refer to relationship terms that appear in a JBN as relationship indicator variables or simply relationship variables. A JBN assigns probabilities to conjunctions of literals of the form $\f_{1}(\bs{\theta_{1}}) = \a_{1}, \ldots, \f_{n}(\bs{\theta_{n}}) =\a_{n}$. We use the term ``Join'' because such conjunctions correspond to table joins in databases. A key step in model learning is to define an {\em empirical distribution} over the random variables in the model. Since the random variables in a JBN are function terms, this amounts to associating a probability with a conjunction $\C$ of literals given a database. We can use the instantiation frequency $\P_{\D}(\C)$ as the empirical distribution over the nodes in a JBN. The goal of JBN structure learning is then to construct a model of $\P_{\D}$ given a database $\D$ as input.

\begin{figure}[htbp] 
   \centering
   \includegraphics[width=3in]{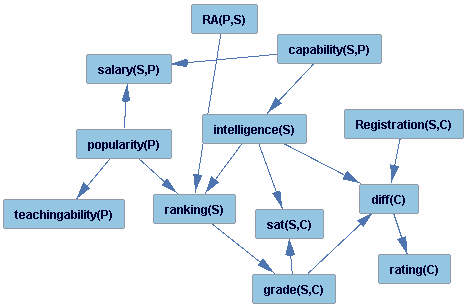} 
   \caption{A Join Bayes Net for the relational schema shown in Table~ref{table:university-schema}.}
   \label{fig:university-JBN}
\end{figure}


\subsection{The Learn-and-Join Algorithm} 

We describe our structure learning algorithm, then discuss its scope and limitations.
The algorithm is based on a general schema for upgrading a propositional BN learner to a statistical relational learner. 
By ``upgrading'' we mean that the propositional learner is used as a function call or module in the body of our algorithm. We require that the propositional learner takes as input, in addition to a single table of cases, also a set of {\em edge constraints} that specify required and forbidden directed edges.
The output of the algorithm is a DAG $\G$ 
for a database $\D$ with variables as specified in Definition ref{def:jbn}. Our approach is to ``learn and join'': we apply the BN learner to single tables and combine the results successively into larger graphs corresponding to larger table joins. 

In principle, a JBN may contain any set of open function terms, depending on the attributes and relationships of interest. 
To keep the description of the structure learning algorithm simple, we assume that a JBN contains a default set of nodes as follows: (1) one node for each descriptive attribute, of both entities and relationships, (2) one Boolean indicator node for each relationship, (3) the nodes contain no constants. For each type of entity, we introduce one first-order variable. 
The algorithm has four phases (pseudocode shown as Algorithm ref{alg:structure}). 

(1) {\em Analyze single tables.} Learn a BN structure for the descriptive attributes of each entity table $\E$ 
of the database separately (with primary keys removed). The aim of this phase is to find within-table dependencies among descriptive attributes (e.g., $\it{intelligence}(\S)$ and $\it{ranking}(\S)$).

(2) {\em Analyze single join tables.} Each relationship table $\R$ is considered. The input table for each relationship $\R$ is the join of that table with the entity tables linked by a foreign key constraint (with primary keys removed). Edges between attributes from the same entity table $\E$ are constrained to agree with the structure learned for $\E$ in phase (1). 
Additional edges from variables corresponding to attributes from different tables may be added. The aim of this phase is to find dependencies between descriptive attributes {\em conditional on} the existence of a relationship.
 This phase also finds dependencies between descriptive attributes of the relationship table $\R$.


(3) {\em Analyze double join tables.} 
The extended input relationship tables from the second phase (joined with entity tables) 
are joined in pairs to form the input tables for the BN  learner. Edges between variables
considered in phases (1) and (2) are constrained to agree with the structures previously learned. The graphs learned for each join pair are merged to produce a DAG $\G$. The aim of this phase is to find dependencies between descriptive attributes {\em conditional on} the existence of a relationship chain of length 2. 

(4) {\em Satisfy slot chain constraints.} For each link $\A \rightarrow \B$ in $\G$, where $\A$ and $\B$ are attributes from different tables, arrows from Boolean relationship variables into $\B$ are added if required to satisfy the following constraints: (1) $\A$ and $\B$ share variable among their arguments, or (2) the parents of $\B$ contain a chain of foreign key links connecting $\A$ and $\B$. 
\begin{figure}[htbp]
\begin{center}
 \includegraphics[width=3in]{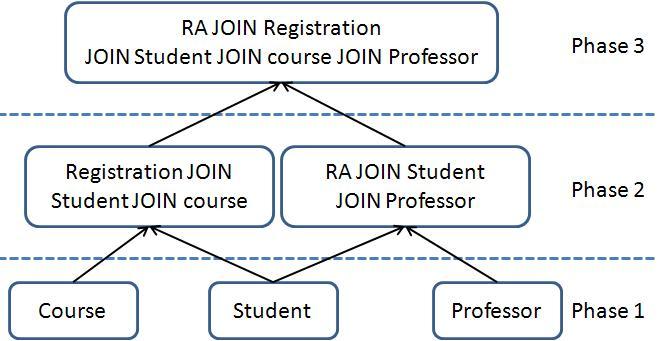} 
\caption{To illustrate the learn-and-join algorithm. A given BN learner is applied to each table and join tables (nodes in the graph shown). The presence or absence of edges at lower levels is inherited at higher levels.}
\label{fig:structure-learn}
\end{center}
\end{figure}
%
Figure~ref{fig:structure-learn} illustrates the increasingly large joins built up in successive phases. Algorithm~ref{alg:structure} gives pseudocode.

%
%
\subsection{Discussion} We discuss the patterns that can be represented in our current JBN implementation and the dependencies that can be found by the learn-and-join algorithm.

\paragraph{Number of Variables for each Entity Type.} There are data patterns that require  more  than one variable per  type to express. For a simple example, suppose we have a social network of friends represented by a $\it{Friend}(\X,\Y)$ relationship both of whose arguments are of type $\it{Person}$. There may be a correlation between the smoking of a person and that  of her friends, so a JBN may place a link between nodes $\it{smokes}(\X)$ and $\it{smokes}(\Y)$, which requires two variables of type $\it{Person}$. In principle, the case of several variables for a given entity type can be translated to the case of one variable per entity type as follows \cite{Schulte08}. For each variable of a given type, make a copy of the entity table. In our example, we would obtain two $\it{Person}$ tables, $\it{Person}_{\X}$ and $\it{Person}_{\Y}$. Each copy can be treated as its own type with just one associated variable. 
We leave for future work an efficient implementation of the learn-and-join algorithm with several variables for a given entity type.

\begin{algorithm}[htb]
\begin{algorithmic}
{\footnotesize
\STATE {\em Input}: Database $\D$ with $E_1,..E_e$ entity tables, $R_1,... R_r$ Relationship tables, 
\STATE {\em Output}: A JBN graph for $\D$. 
\STATE {\em Calls}: PBN: Any propositional Bayes net learner that accepts edge constraints and a single table of cases as input. 
\STATE {\em Notation}: PBN$(\T,\mbox{Econstraints})$ denotes the output DAG of PBN. Get-Constraints$(\G)$ specifies a new set of edge constraints, namely that all edges in $\G$ are required, and edges missing between variables in $\G$ are forbidden.
} 
\end{algorithmic}
\begin{algorithmic}[1]
{\footnotesize
	\STATE Add descriptive attributes of all entity and relationship tables as variables to  $G$. Add a boolean indicator for each relationship table to $G$.
	\STATE Econstraints = $\emptyset$ {[Required and Forbidden edges]} 
\FOR {m=1 to e}
	\STATE Econstraints += Get-Constraints(PBN($E_m$ , $\emptyset$)) 
	\ENDFOR	
\FOR {m=1 to r}
	\STATE $N_m$ :=  
	join of $R_m$ and entity tables linked to $R_m$ 
	\STATE Econstraints += Get-Constraints(PBN($N_m$, Econstraints))
\ENDFOR
\FORALL{$N_i$ and $N_j$ with an entity table foreign key in common}
	\STATE $K_{ij}$ :=  
	join of $N_i$ and $N_j$ 
	\STATE Econstraints += Get-Constraints(PBN($K_{ij}$, Econstraints))
\ENDFOR
\STATE Construct DAG $\G$ from Econstraints
\STATE Add edges from Boolean relationship variables to satisfy slot chain constraints
\STATE Return $\G$
		} 
\end{algorithmic}
\caption{Pseudocode for structure learning \label{alg:structure}}
\end{algorithm}

\paragraph{Correlation Coverage.} 
The join-and-learn algorithm finds correlations between descriptive attributes, within a single table and between attributes from different linked tables. 
It does not, however, find dependencies between relationship variables (e.g., $\it{Married}(\X,\Y)$ predicts $\it{Friend}(\X,\Y)$). 
A JBN search for such dependencies could use local search methods such as those described  in \cite{Kersting2007,Getoor2007c}.
In sum, the learn-and-join algorithm is suitable when the goal is to find correlations between descriptive attributes conditional on a given link structure.

\section{Parameter Learning in Join Bayes Nets} \label{sec:mle}

This section treats the problem of computing conditional frequencies in the database distribution, which corresponds to computing sample frequencies in the single table case. 
The main problem is computing probabilities conditional on the {\em absence} of a relationship. 
This problem arises because a JBN includes relationship indicator variables such as $reg(\S,\course)$, and building a JBN therefore requires modelling the case where a relationship does not hold. We apply the recent virtual join (VJ) algorithm of \cite{Khosravi2009} to address this computational bottleneck. The key constraint that the VJ algorithm seeks to satisfy is to {\em avoid enumerating the number of tuples that satisfy a negative relationship literal.} 
A numerical example illustrates why this is necessary. Consider a university database with 20,000 Students, 1,000 Courses and 2,000 TAs. If each student is registered in 10 courses, the size of a $\it{Registered}$ table is 200,000, or in our notation $\grounds_{\D}(\it{Registered}(\S,\course))_{\D}) = 2 \times 10^{5}$. So the number of complementary student-course pairs is $\grounds_{\D}(\neg \it{Registered}(\S,\course))) = 2 \times 10^{7}-2 \times 10^{5}$, which is a much larger number that is too big for most database systems. 
If we consider joins, complemented tables are even more difficult to deal with: suppose that each course has at most 3 TAs. Then  $\grounds_{\D}(\it{Registered}(\S,\course),\it{TA}(\T,\course)) < 6  \times 10^{5}$, whereas $\grounds_{\D}(\neg \it{Registered}(\S,\course),\neg \it{TA}(\T,\course))$ is on the order of $4 \times 10^{10}$. 
After explaining the basic idea behind the VJ algorithm, we give pseudocode for utilizing it in to estimate CP-table entries via joint probabilities. We conclude with a run-time analysis.

\subsection*{The Virtual Join algorithm for CP-tables.} 
Instead of computing conditional frequencies, the VJ algorithm computes joint probabilities of the form $P_{\D}(\it{child},\it{parent\_values})$, which correspond to conjunctions of literals. Conditional probabilities are easily obtained from the joint probabilities by summation. 
While generally it is easier to find conditional rather than joint probabilities, there is an efficient dynamic programming scheme 
(informally a ``1 minus'' trick)
 that relies on the simpler structure of conjunctions. 
%
The enumeration of groundings for negative literals can be avoided by recursively applying a probabilistic principle (a ``1-minus'' scheme). Consider the equation
$P(\C) = P(\C,\L) + P(\C, \neg \L)$, which entails
\begin{eqnarray}
P(\C, \neg \L) = P(\C) - P(\C, \L).	
	\label{eq:dynamic}
\end{eqnarray}

where $\C$ is a conjunction of literals and $\L$ a literal. This equation shows that the  computation of a probability involving a negative relationship literal $\neg \L$ can be recursively reduced to two computations, one with the positive literal $\L$ and one with neither $\L$ nor $\neg \L$, each of which contains one less negative relationship literal. 
In the base case, all literals are positive, so the problem is to find $P_{\D}(\C)$ for database instance $\D$ where $\C$ contains positive relationship literals only. 
This can be done with a standard database table join.
Figure ref{fig:example} illustrates the recursion.
\begin{figure*}[htbp] 
   \centering
   \includegraphics[width=6in]{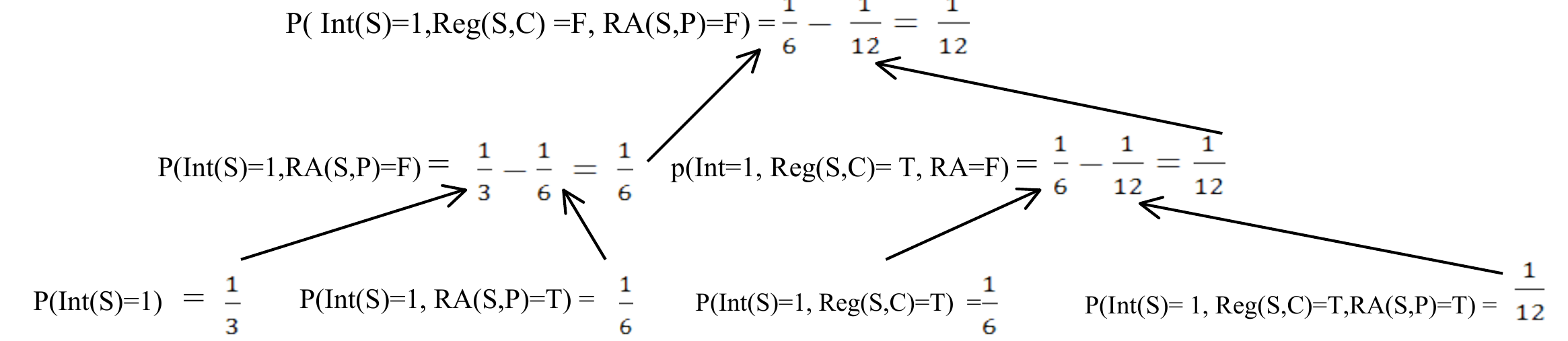} 
   \caption{A frequency with 
   negated relationships (nonexistent links) can be computed from frequencies that involve positive relationships (existing links) only. The leaves in the computation tree involves existing database tables only.
   The subtractions involve looking up results of previous computations. 
To reduce clutter, we abbreviated some of the predicates.}
   \label{fig:example}
\end{figure*}
%

The VJ algorithm computes the database frequency for just a single input conjunction of literals. We adapt it for CP-table estimation with two changes: (1) We compute all frequencies that are defined by the same join table at once when the join has been built, and (2) we use the CP-table itself as our data structure for storing the results of intermediate computations from which other database frequencies are derived. The algorithm can be visualized as a dynamic program that successively fills in rows in a \textbf{joint probability table}, or \textbf{JP-table}, where we first fill in rows with 0 nonexistent relationships, then rows with 1 nonexistent relationship, etc. A JP-table is just like a CP-table whose rows correspond to joint probabilities rather than conditional probabilities. Algorithm~ref{alg:adapted} shows the pseudocode for the VJ algorithm.

\begin{algorithm}
\begin{algorithmic}
\STATE \underline{Notation}: A row $r$ corresponds to a partial or complete assignment for function terms. The value assigned to function term $\f(\set{\theta})$ in row $r$ is denoted by $\f_{r}$. The probability for row $r$ stored in JP-table $\tau$ is denoted by $\tau(r)$. 
\STATE \underline{Input}: database $\D$; child variable and parent variables divided into a set $\set{\R_{1},\ldots,\R_{m}}$ of relationship predicates and a set $\set{C}$ of function terms that are not relationship predicates.
 \STATE \underline{Calls}: initialization function $\join(\set{\C},\true = \R_1 = \cdots = \R_k)$. Computes join frequencies conditional on relationships $\R_{1},\ldots,\R_{k}$ being true.
\STATE \underline{Output}: Joint Probability table $\tau$ such that the entry $\tau(r)$ for row $r \equiv
(\set{\C} = \set{\C}_{r}, \set{\R} = \set{\R}_{r})$ is the frequencies $P_{\D}(\set{\C} = \set{\C}_{r}, \set{\R} = \set{\R}_{r})$ in the database distribution $\D$.
\end{algorithmic}
\begin{algorithmic}[1]

\STATE \COMMENT{fill in rows with no false relationships using table joins}\label{line:start-join}
\FORALL{valid rows $r$ with no assignments of $\false$ to relationship predicates}
\FOR{$i=0$ to $m$}
\IF[$r$ has $m-i$ unspecified relationship literals]{$r$ has exactly $i$ true relationship literals $\R^1,..,\R^i$}
\STATE find $P_{\D}(\set{\C} =\set{\C}_{r}, \set{\R} = \set{\R}_{r})$ using $\join(\set{\C} =\set{\C}_{r}, \set{\R} = \set{\R}_{r})$. Store the result in $\tau(r)$. \label{line:join}
\ENDIF
\ENDFOR
\ENDFOR \label{line:end-join}

\STATE \COMMENT{Recursively extend the table to JP-table entries with false relationships.}
\FORALL{
rows $r$ with at least one assignment of $\false$ to a relationship predicate}
\FOR{$i=1$ to $m-1$}
\IF[find conditional probabilities when $\R^1$ is true and when unspecified]
{$r$ has exactly $i$ false relationship literals $\R^1,..,\R^i$}
\STATE Let $r_{\true}$ be the 
row such that (1) $\R^1_{r_{\true}} = \true$, (2) $\f^{R^1}_{r_{\true}}$ is unspecified for all attributes $\f^{R^1}$ of $\R^1$, and (3) $r_{\true}$ matches $r$ on all other variables. 
\STATE Let $r_{*}$ be the 
row that matches $r$ on all variables $\f$ that are not $\R^1$ or an attribute of $\R^1$ and has $\R^1_{r_{*}}$ unspecified. 
\STATE \COMMENT{The rows $r_{*}$ and $r_{\true}$ have one less false relationship variable than $r$.}
\STATE Set $\tau(r) := \tau(r_{*}) - \tau(r_{\true}).$ \label{line:update}
\ENDIF
\ENDFOR
\ENDFOR
\end{algorithmic}
 \caption{A dynamic program for estimating JP-table entries in a Join Bayes Net.\label{alg:adapted}}
\end{algorithm}

\subsection*{Implementation and Complexity Analysis.} The intermediate results of the computation are stored in an extended JP-table structure that features a third value $*$ for relationship predicates (in addition to $\true,\false$). This is used to represent the frequencies where a relationship predicate and its attributes are unspecified.  

The algorithm satisfies the key constraint that it never enumerates the groundings that satisfy a negative relationship literal. A detailed complexity analysis of the VJ algorithm is given in~\cite{Khosravi2009}; we summarize the main points that are relevant for CP-table estimation. 
Essentially, each computation step fills in one entry in the extended JP-table. 
Compared to the CP-tables for a given Join Bayes net structure, our algorithm adds an extra auxilliary value * to the domain of each relationship indicator variable.
Thus the increase in the size of the data structure is manageable compared to the CP-table in the original JBN. An important point is that the recursive update in Line~ref{line:update} does not require a database access. Data access occurs only in Lines~ref{line:start-join}--~ref{line:end-join}. Therefore the cost in terms of database accesses is essentially the cost of counting frequencies in a join table comprising all $m$ relationships that occur in the child or parent nodes (Line~ref{line:join} with $i=m$). This computation is necessary for any algorithm that estimates the CP-table entries. It can be optimized using the tuple ID propagation method of~\cite{yin2008,Yin2004}. 
The crucial parameter for the complexity of this join is the number $m$ of relationship predicates. In our learn-and-join algorithm, this parameter is bounded at $m=2$. \cite{Khosravi2009} provide an analysis whose upshot is that for typical SRL applications this parameter can be treated as a small constant. A brief summary of the reasons is as follows. (1) The space of models or rules that need to be searched becomes infeasible if too many relationships are considered at once. (2) Patterns that involve many relationships, for instance relationship chains of length 3 or more, are hard to understand for users. (3) Objects related by long relationship chains tend to carry less statistical information about a target object. 
We next examine empirically the performance of our our structure and parameter
learning algorithms.

\section{Empirical Evaluation of JBN Learning Algorithms}
The learn-and-join algorithm trades off learning complexity with the coverage of data correlations. In this section we evaluate both sides of this trade-off on three datasets: the run-time of the algorithm, especially as the dataset grows larger, and its performance in predicting database probabilities at the class level, which is the main task that motivates our algorithm. The more important dependencies the algorithm misses, the worse its predictive performance, so evaluating the predictions provides information about the quality of the JBN structure learned.

\subsection{Implementation and Datasets}

  All experiments were done on a QUAD CPU Q6700 with a 2.66GHz CPU and 8GB of RAM. The implementation used many of the procedures in version 4.3.9-0 of CMU's Tetrad package \cite{2008a}. For single table BN search we used the Tetrad implementation of GES search \cite{Chickering2002} with the BDeu score (structure prior uniform, ESS=8). Edge constrains were implemented using Tetrad's ``knowledge'' functionality. JBN inference was carried out with Tetrad's Rowsum Exact Updater algorithm. Our datasets and code are available for ftp download from ftp://ftp.fas.sfu.ca/pub/cs/oschulte.
 
\subsection*{Datasets}

{\em University Database.} In order to check the correctness of our algorithms directly, we manually created a small dataset, based on the schema given in ref{table:university-schema}. 
The entity tables contain 38 students, 10 courses, and 6  Professors. The $reg$ table has 92 rows and the $\it{RA}$ table has 25 rows. 



{\em MovieLens Database.} The second dataset is the MovieLens dataset from the UC Irvine machine learning repository. 
It contains two entity tables: $\it{User}$ with 941 tuples and $\it{Item}$ with 1,682 tuples, and one relationship table $\it{Rated}$ with 80,000 ratings. The $\it{User}$ table has 
3 descriptive attributes $\age, \it{gender}, \it{occupation}$. We discretized the attribute age into three bins with equal frequency. The table $\it{Item}$ represents information about the movies. It has 17 Boolean attributes that indicate the genres of a given movie; a movie may belong to several genres at the same time. For example, a movie may have the value $\true$ for both the $\it{war}$ and the $\it{action}$ attributes. We performed a preliminary data analysis and omitted genres that have only weak correlations with the rating or user attributes, leaving a total of three genres. 

{\em Financial Database.} The third dataset is a modified version of the financial dataset from the PKDD 1999 cup. 
There are two entity tables, $\it{Client}$ with 5,369 tuples and $\it{Account}$ with 4,500 tuples. Two relationship tables, $\it{CreditCard}$ with 2,676 tuples and $\it{Disposition}$ with 5,369 tuples relate  a client with an account. 
The Client table has 10 descriptive attributes: the client's age, gender and 8 attributes on demographic data of the client. The Account table has 3 descriptive attributes: information on loan amount associated with an account, account opening date, and how frequently the account is used.

\subsection{Learning: Experimental Design}

To benchmark the runtimes of our learning algorithm, we applied the structure learning routine \texttt{learnstruct} (default options) of the Alchemy package for MLNs \cite{Domingos07}. We chose MLNs for the following reasons: (1) MLNs are currently one of the most active areas of SRL research (e.g., \cite{Domingos07,Huynh2008}). Part of the reason for this is that undirected graphical models avoid the computational and representational problems caused by cycles in instance-level directed models (Section~ref{sec:related}). Discriminative MLNs can be viewed as logic-based templates for conditional Markov random fields, a prominent formalism for relational classification \cite{Taskar2002}. 
(2) Alchemy provides open-source, state-of-the-art learning software for MLNs. Structure learning software for alternative systems like BLNs and PRMs is not easily available.\footnote{The webpage \cite{bib:kersting-page} lists different SRL systems and which software is available for them. The Balios BLN engine \cite{Kersting2007} supports only parameter learning, not structure learning; see \cite{bib:kersting-page}. We could not obtain source code for PRM structure learning.} (3) BLNs and PRMs require the specification of additional structure like aggregation functions or combining rules. This confounds our experiments with more parameters to specify. Also, incorporating the extra structure complicates learning for these formalisms, so arguably a direct comparison with JBN learning is not fair. In contrast, the MLN formalism, like the JBN, does not require extra structure beyond the relational schema. 




\subsection{Learning: Results}
We present results of applying our learning algorithms to the three relational datasets. The resulting JBNs are shown in Figures ref{fig:university-JBN}, ref{fig:structmovie}, and ref{fig:structfinancial}.  In the MovieLens dataset, the algorithm finds a number of cross-entity table links involving the age of a user. Because genres have a high negative correlation, the algorithm produces a dense graph among the genre attributes.  The richer relational structure of the Financial dataset is reflected in a more complex graph with several cross-table links. The birthday of a customer (translated into discrete age levels) has especially many links with other variables.
The CP-tables for the learned graph structures  were filled in using the dynamic programming algorithm~ref{alg:adapted}. Table ref{table:runtime} presents a summary of the run times for the datasets.

The databases translate into ground atoms for Alchemy input as follows: University 390, MovieLens 39,394, and Financial 16,129. On our system, Alchemy was able to process the  University database, but did not have sufficient computational resources to return a result for the MovieLens and Financial data. 
We therefore subsampled the datasets to obtain small databases on which we can compare Alchemy's runtime with that of the join-and-learn algorithm. Because Alchemy returned no result on the complete datasets, we formed three subdatabases by randomly selecting entities for each dataset. We restricted the relationship tuples in each subdatabase to those that involve only the selected entities. The resulting subdatabase sizes are as follows. For MovieLens, (i) 100 users, 100 movies = 1,477  atoms (ii) 300 users, 300 movies = 9698  atoms (iii) 500 users, 400 movies = 19,053  atoms. For Financial, (i) 100 Clients, 100 Accounts = 3,228  atoms, (ii) 300 Clients, 300 Accounts = 9,466  atoms, (iii) 500 Clients, 500 Accounts = 15,592 atoms.  For the Financial dataset, which contains numerous descriptive attributes, Alchemy returned a result only for the smallest subdatabase (i). Table~ref{table:runtime} shows that the runtime of the JBN learning algorithm applied to the entire dataset is 600 times faster than Alchemy's learning time on a dataset about half the size. We emphasize that this is not a criticism of Alchemy structure learning, which aims to find  a structure that is optimal for instance-level predictions (cf. Section~ref{sec:related}). Rather, it illustrates that the task of finding class-level dependencies is much less computationally complex taken as an independent task 
than when it is taken in conjunction with optimizing instance-level predictions. The next section compares the probabilities estimated by the JBN with the database frequencies computed directly from SQL queries. 

\begin{figure}[h] 
   \centering
   \includegraphics[scale=0.5]{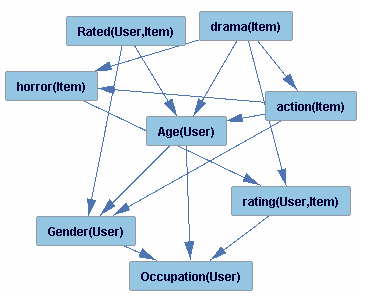} 
   \caption{The JBN structure learned by our merge learning algorithm~ref{alg:structure} for the MovieLens Dataset.}
   \label{fig:structmovie}
\end{figure}
\begin{figure}[h] 
   \centering
   \includegraphics[scale=0.4]{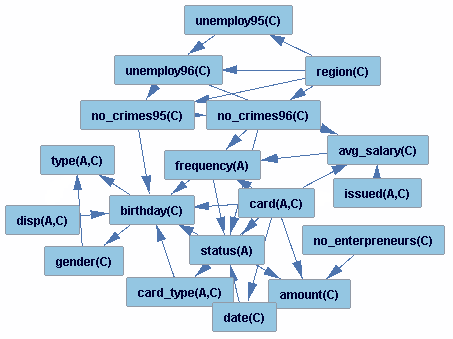} 
   \caption{The JBN structure learned by our merge learning algorithm~ref{alg:structure} for the Financial Dataset.}
   \label{fig:structfinancial}
\end{figure}




\begin{table}[htb]
\centering
\small
\label{note}
\begin{tabular}
{|l|l|l|l|l|l|} \hline
{\bf Dataset} & \multicolumn{2}{c|}{\textbf{JBN}} & \multicolumn{3}{c|}{\textbf{MLN-subdatasets}} \\ \hline
& {\bf PL} & {\bf SL}&  {\bf small}& {\bf medium}& {\bf large}\\ \hline
University &0.495 & 0.64 & & &90  \\ \hline
Movie Lens & 35 & 15 & 455& 1,316&33,656 \\ \hline
Financial & 136 & 52 & 21,607 & n/a & n/a\\ \hline
\end{tabular}
\caption{The runtimes---in seconds---for structure learning (SL) and parameter learning (PL) on our three datasets.}
\label{table:runtime}
\end{table}

\subsection{Inference: Experimental Design}
A direct comparison of class-level inference with other SRL formalisms is difficult as the available implementations support only instance-level queries.\footnote{For example, Alchemy and the Balios BLN engine \cite{Kersting2007} support only queries with ground atoms as evidence. We could not obtain source code for PRM inference.} This reflects the fact that current SRL systems are designed for the task of predicting attributes of individual entities, rather than class-level prediction. We therefore evaluated the class-level predictions of our system against gold standard frequency counts obtained directly from the data using SQL queries. 
We follow the approach of \cite{Getoor2001} in our experiments and use the sample frequencies for parameter estimation. This is appropriate when the goal is to evaluate whether the statistical model adequately summarizes the data distribution. If the data frequencies are smoothed to support generalizations beyond the data, the first step is to compute the data frequencies, so our experiments are relevant to this case too.

We randomly generated 10 queries for each data set according to the following procedure. First, choose randomly a target node $\V$ and a value $a$ such that $P(\V=a)$ is the probability to be predicted. Then choose randomly the number $k$ of conditioning variables, ranging from 1 to 3. Make a random selection of $k$ variables $\V_{1},\ldots,\V_{k}$ and corresponding values $a_{1},\ldots,a_{k}$. The query to be answered is then $P(\V=a|\V_{1} = a_{1},\ldots,\V_{k}= a_{k})$. Table~ref{table:inference} shows representative test queries. It compares the probabilities predicted by the JBN with the frequencies in the database as computed by an SQL query, as well as the runtimes for computing the probability using the JBN vs. the SQL.
\begin{table*}
\centering
{\footnotesize \begin{tabular}
{|l|l|l|l|l|} \hline
\textit{Dataset}&{\bf Query} & {\bf P(Model)/time} & {\bf P(SQL/time)} \\ \hline
University & p($ranking = 3| RA = \false, \textit{popularity} = 1, \intelligence = 2$) &0.6048/ 2.12 & 0.6086/ 0.06  \\ \hline
University &p($grade = 3| registration = \true, \textit{popularity} = 1, \intelligence = 2$) & 0.1924/2.03 & 0.2/0.064 \\ \hline
MovieLens & p($Age = 2| Rated = \true, rating = 4, horror = 1$) & 0.0878/0.78 & 0.09 /12.23  \\ \hline
MovieLens& p($Age = 1| Rated = \false, rating = 4, horror = 1$) & 0.5433/ 1.23 & 0.5166/14.34  \\ \hline
MovieLens& p($Gender = m| Rated = \false, Age = 1, horror = 1 , action = 1 $) & 0.6904/0.97 & 0.6857/9.66  \\ \hline
MovieLens& p($horror = 1| Rated = \true, Age = 1, Gender = F , action = 1 $) & 0.0433/1.20 & 0.0293/7.23  \\ \hline
Financial& p($Gender = M| date = 1, status = A$) & 0.5211/11.32 & 0.5391/6.37  \\ \hline
Financial& p($Gender = M| date = 1, status = A, Disp = \false$) & 0.5187/10.65 & No result \\ \hline
\end{tabular}
}
\caption{The table shows representative randomly generated queries. We compare the probability estimated by our learned JBN model (P(Model)) with the database frequency computed from direct SQL queries (P(SQL)), and the execution times (in seconds) for each inference method. The SQL query in the last line exceeded our system resources. For simplicity we omitted first-order variables.}
\label{table:inference}
\end{table*}
%

\subsection{Inference: Results} 
The averages reported are taken over 10 random queries for each dataset. We see in Table~ref{table:inference} and Figure~ref{fig:results} that the predicted probabilities are very close to the data frequencies: The average difference is less than 3\% for MovieLens, and less than 8\% for Financial. The measurements for Financial are taken on queries with positive relationship literals only, because the
SQL queries that involve negated relationships did not terminate with a result  for the Financial dataset. The graph shows the nontermination as corresponding to a high time-out number. 

Observations about processing speed that hold for all datasets include the following. (1) For queries involving {\em negated relationships}, JBN inference was much faster on the MovieLens dataset. SQL queries with negated relationships were infeasible on the Financial dataset, whereas the JBN returns an answer in around 10 seconds. (2) Both JBN and SQL queries speed up as the number of conditioning variables increases. (3) Higher probability queries are slower with SQL, because they correspond to larger joins, whereas the size of the probability does not affect JBN query processing. 
\begin{figure}[htb]
   \centering
   \includegraphics[width=3.3in]{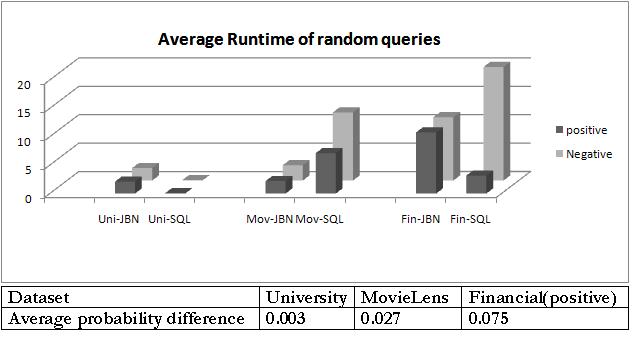} 
   \caption{Comparing the probability estimates and times (in sec) for obtaining them from the learned JBN models versus SQL queries.}
   \label{fig:results}
\end{figure}

Other observations depend on the difference between the datasets: MovieLens has relatively many tuples and few attributes, whereas Financial has relatively few tuples and many attributes. These factors differentially affected the performance of SQL vs. JBN inference as follows (for queries involving positive relationships only). 
(1) A larger number of attributes and/or a larger number of categories in the attributes in the schema decreases the speed of both JBN inference and SQL queries. But the slowdown is greater for JBN queries because the required marginalization steps are expensive (summing over possible values of nodes). (2) The number of tuples in the database table is a very significant factor for the speed of SQL queries but does not affect JBN inference. This last point is an important observation about the {\em data scalability of JBN inference}: While the cost of the learning algorithms depends on the size of the database, once the learning is completed, query processing is independent of database size. So for applications like query optimization that involve many calls to the statistical inference procedure, the investment in learning a JBN model is quickly amortized in the fast inference time.

\section{Conclusion}
Class-level generic dependencies between attributes of linked objects and of links are important in themselves, and they support applications like policy making, strategic planning, and query optimization. The focus on class-level dependencies brings gains in tractability of learning and inference. The theoretical foundation of our approach is classic AI research which established a definition of the frequency of a first-order formula in a given relational database. We described efficient and scalable algorithms for structure learning and parameter estimation in Join Bayes nets, which model the database frequencies over attributes of linked objects and links. Our algorithms upgrade a single-table Bayes net learner as a self-contained module to perform relational learning.  JBN inference can be carried out with standard algorithms ``as is'' to answer class-level probabilistic queries. An evaluation of our methods on three data sets shows that they are computationally feasible for realistic table sizes, and that the learned structures represented the statistical information in the databases well. Querying database statistics via the net is often faster than directly with SQL queries, and does not depend on the size of the database. 

A fundamental limitation of our approach is that Join Bayes nets do not directly support instance-level queries (our model does not include a solution to the combining problem). 
 Limitations that can be addressed in future research include restrictions on the types of correlation that our structure learning algorithm can discover, such as dependencies between relationships, and dependencies that require more than one first-order variable per entity type to represent.

\section*{Acknowledgments} This research was supported by Discovery Grants to the first and fifth author from the Natural Sciences and Engineering Research Council of Canada. A preliminary version of our results was presented at the %
STRUCK and GKR workshops at IJCAI 2009, and at the Computational Intelligence Forum at the University of British Columbia. 
We thank the audiences and Ke Wang for helpful comments.

\bibliographystyle{abbrv}
\bibliography{master}
\end{document}